\documentclass[letterpaper, 10 pt, conference]{formatting/ieeeconf}
\IEEEoverridecommandlockouts                              
\let\labelindent\relax
\usepackage{amsfonts}       

\usepackage{amsthm}
\usepackage{mathtools}      
\usepackage{amssymb}        
\usepackage{filecontents}
\usepackage{graphicx}       
\usepackage{marginnote}     
\usepackage{marvosym}       
\usepackage{overpic}        
\usepackage{tabularx}
\usepackage{cite}
\usepackage{color}
\usepackage[linesnumbered,algoruled,boxed,lined,noend]{algorithm2e}
\usepackage[normalem]{ulem}
\usepackage{enumitem}
\usepackage[normalem]{ulem}

\usepackage{lipsum}

\usepackage{comment}
\usepackage[export]{adjustbox}
\usepackage{epstopdf}

\usepackage{url}
\usepackage[font=small]{caption}
\usepackage{subcaption}
\newif\ifdraft
\draftfalse

\ifdraft
\usepackage[paperheight=11in,paperwidth=9.5in,
			left=1.25in,right=1.25in,
			top=0.75in,bottom=0.75in,
			heightrounded,marginparwidth=1.2in,
			marginparsep=0.05in]{geometry}
\usepackage{xcolor}
\usepackage{xargs} 
\usepackage[textsize=footnotesize]{todonotes}
\newcommandx{\nt}[2][1=]{\todo[linecolor=red,
			backgroundcolor=red!10,bordercolor=red,#1]{#2}}
\newcommandx{\jy}[2][1=]{\todo[linecolor=green,
			backgroundcolor=green!10,bordercolor=green,#1]{JY:#2}}
\newcommandx{\kg}[2][1=]{\todo[linecolor=green,
			backgroundcolor=green!10,bordercolor=blue,#1]{KG:#2}}
\newcommandx{\sw}[2][1=]{\todo[linecolor=blue,
			backgroundcolor=orange!10,bordercolor=orange,#1]{SW:#2}}
\newcommandx{\zy}[2][1=]{\todo[linecolor=purple,
			backgroundcolor=purple!10,bordercolor=purple,#1]{ZY:#2}}
\newcommandx{\dz}[2][1=]{\todo[linecolor=pink,
			backgroundcolor=pink!10,bordercolor=pink,#1]{dz:#2}}

\else
\usepackage{xcolor}
\usepackage{xargs} 
\usepackage[textsize=footnotesize]{todonotes}
\newcommandx{\nt}[2][1=]{\todo[linecolor=red,
			backgroundcolor=red!10,bordercolor=red,#1]{#2}}
\newcommandx{\jy}[2][1=]{\todo[linecolor=green,
			backgroundcolor=green!10,bordercolor=green,#1]{JY:#2}}
\newcommandx{\kg}[2][1=]{\todo[linecolor=green,
			backgroundcolor=green!10,bordercolor=blue,#1]{KG:#2}}
\newcommandx{\sw}[2][1=]{\todo[linecolor=blue,
			backgroundcolor=orange!10,bordercolor=orange,#1]{SW:#2}}
\newcommandx{\zy}[2][1=]{\todo[linecolor=purple,
			backgroundcolor=purple!10,bordercolor=purple,#1]{ZY:#2}}
\newcommandx{\dz}[2][1=]{\todo[linecolor=pink,
			backgroundcolor=pink!10,bordercolor=pink,#1]{dz:#2}}
\fi

\newif\iftwocolumn
\twocolumntrue

\theoremstyle{definition}

\theoremstyle{remark}

\SetKwProg{Fn}{Function}{}{}
\SetKwComment{Comment}{$\triangleright$\ }{}

\makeatletter
\def\subsubsection{\@startsection{subsubsection}
                                 {3}
                                 {\z@ \hspace*{1mm}}
                                 {0ex plus 0.1ex minus 0.1ex}
                                 {0ex}
                                 {\normalfont\normalsize\itshape}}
\makeatother

\def\toro{\texttt{TORO}\xspace}

\def\bl#1{\textcolor{blue}{#1}}
\def\ours{\texttt{MODAP}\xspace} 

\font\titlefont=ptmb at 14.3pt
\title{\titlefont
Toward Holistic Planning and Control Optimization for Dual-Arm Rearrangement  
}
\author{
Kai Gao$^*$ \qquad Zihe Ye$^*$ \qquad Duo Zhang$^*$  \qquad Baichuan Huang \qquad Jingjin Yu
\thanks{
The authors are with the Department of Computer Science, Rutgers, the State University of New Jersey, Piscataway, NJ, 
USA. 
$^*$These authors made equal contributions to the study.
}%
}

\begin{document}

\maketitle
\thispagestyle{empty}
\pagestyle{empty}


\begin{abstract}
Long-horizon task and motion planning (TAMP) is notoriously difficult to solve, let alone optimally, due to the tight coupling between the interleaved (discrete) task and (continuous) motion planning phases, where each phase on its own is frequently an NP-hard or even PSPACE-hard computational challenge. 
In this study, we tackle the even more challenging goal of jointly optimizing task and motion plans for a real dual-arm system in which the two arms operate in close vicinity to solve highly constrained tabletop multi-object rearrangement problems. 
Toward that, we construct a tightly integrated planning and control optimization pipeline, \bl{M}akespan-\bl{O}ptimized \bl{D}ual-\bl{A}rm \bl{P}lanner (\ours) that combines novel sampling techniques for task planning with state-of-the-art trajectory optimization techniques. 
Compared to previous state-of-the-art, \ours produces task and motion plans that better coordinate a dual-arm system, delivering significantly improved execution time improvements while simultaneously ensuring that the resulting time-parameterized trajectory conforms to specified acceleration and jerk limits. 
\end{abstract}

\section{Introduction}\label{sec:intro}
An ultimate goal in intelligent robotics (and, more generally, embodied AI) research and development is to enable robots/agents to plan and execute long-horizon tasks autonomously. We would like robots to effectively tackle \emph{task and motion planning} (TAMP) challenges and deliver solutions that are as optimal as possible. However, TAMP challenges are generally computationally intractable to solve optimally, even at the sub-problem level. Take tabletop object rearrangement as an example, the task planning portion is NP-hard to solve optimally for multiple natural objectives, including minimizing the total number of pick-n-place operations \cite{han2018complexity} and minimizing the number of temporarily dislocated objects \cite{gaorunning}. On the other hand, the computational complexity of motion planning has long been established to be at least PSPACE-hard \cite{hopcroft1984complexity}, with real-world computational cost growing exponentially with respect to the number of dimensions. We note this is merely at the path planning level without considering robot dynamics. Yet, actual path executions on robots must be dynamically feasible and as optimal as possible, creating further computational requirements.   
Despite the computational obstacles involved, given its importance, research efforts on better solving TAMP have continued to pick up pace in recent years \cite{zhang2022visual, wang2022hierarchical, kim2022representation}. 

\begin{figure}[h]
    \centering
\includegraphics[width=0.5\textwidth]{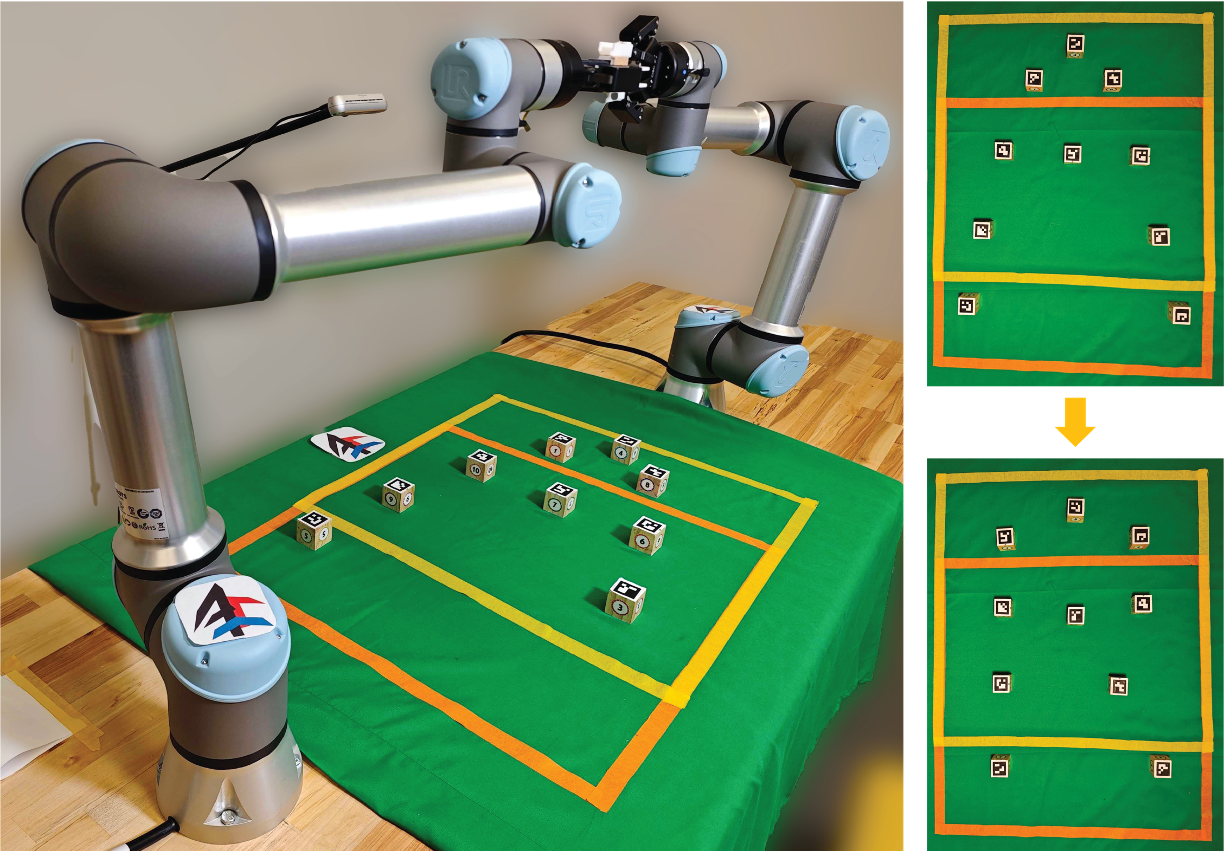}
    \caption{The experimental setup used in the study contains two Universal Robots UR-5e arms, each equipped with a Robotiq 2F-85 two-finger gripper. [left] A handoff operation during the plan execution to convert an object layout from the letter 'A' to the letter 'R'. [top right] The start object layout forming the letter 'A', [bottom right] The goal object layout showing the letter 'R'.}
    \label{fig:handoff}
        \vspace{-5mm}
\end{figure}

This work addresses a challenge at the forefront of TAMP research: developing an integrated planning and control optimization framework to enable high-DOF robot systems to solve complex tasks as quickly as possible. 
The key focus of our study is the joint optimization of task planning, path planning, and motion planning to satisfy robots' dynamics constraints.  
Because task and path planning are computationally intractable individually, computing a truly optimal solution is out of the question. Therefore, our work centers on fusing principled methods and heuristics to deliver the best solution.
As an application, we consider the \emph{cooperative dual-arm rearrangement} (CDR) problem \cite{gao2022toward}. Specifically, we solve tabletop object rearrangement (or \toro \cite{han2018complexity}) tasks using a dual-arm manipulator setup. In the setup shown in Fig. \ref{fig:handoff}, each manipulator is a 7-DOF system (a 6-DOF robot arm $+$ 1-DOF parallel 2-finger gripper) with a fixed base. The two manipulators are tasked to solve \toro instances collaboratively, working in close vicinity. 

Our algorithmic solution, \emph{Makespan-Optimized Dual-Arm (Task and Motion) Planner} (\ours), builds on our previous structural and algorithm study of CDR \cite{gao2022toward}, in which several effective search-based solutions that \emph{compacts} an initial plan for CDR were developed. \ours significantly expands over \cite{gao2022toward} with the following main contributions: 
\begin{itemize}[leftmargin=4mm]
    \item \ours intelligently samples inverse kinematics (IK) solutions during the task planning (i.e., when objects are picked or placed) phase, generating multiple feasible motion trajectories for downstream planning and improving long-horizon plan optimality. In contrast, only a shallow plan compactification approach was performed in \cite{gao2022toward}.
    \item \ours leverages a recent GPU-accelerated planning framework,  cuRobo \cite{cuRobo}, to quickly generate high-quality motion plans for our dual-arm system, resulting in magnitudes faster computation of dynamically smooth motions.\footnote{We note that, while cuRobo can be applied to dual-arm motion planning, it cannot be directly applied to solve CDR because the two robots' plans cannot be readily synchronized due to asynchronous pick-n-place actions.} \ours performs further trajectory optimization over the trajectories from cuRobo to accelerate the full plan.
\end{itemize}
Extensive simulation studies and real-robot experiments show that \ours generates dynamically feasible trajectories and is up to $40\%$ more time efficient than the baseline solution (an optimized version of the best method from \cite{gao2022toward}), particularly in highly constrained scenarios.

\section{Related Work}\label{sec:related}
\textbf{Task and Motion Planning.} 
Human activities are generally driven by Task and Motion Planning (TAMP), which provides a natural model for robot planning. 
In recent years, integrated Task and Motion Planning (TAMP) \cite{garrett2021review} has been extensively studied, yielding significant advancements. 
\cite{toussaint2015logic} proposes logic-geometric programming, resulting in an optimization-based approach, to solve combined task and motion planning. \cite{dantam2016incremental,dantam2018incremental} tackles the challenges via incrementally incorporating additional planning constraints. \cite{kaelbling2011hierarchical} approaches integrated TAMP through a hierarchical planning framework. In that regard,  \cite{wang2022hierarchical} investigates how to generate hierarchical policies for grasping tasks in cluttered scenes with latent plans. 
Lately, \cite{luo2024Multistage} also utilizes hierarchical TAMP to resolve a multi-stage cable routing problem. \cite{wells2019learning} improves the scalability of TAMP in tabletop scenarios in the setup with one fixed robot by introducing geometric knowledge into the task planner. As task plan may require an ordered description, \cite{garrett2020pddlstream} defined PDDL (Planning Domain Definition Language) for task planning description, and \cite{zhao2023llm} utilizes LLM (large language model) to encode commonsense knowledge for large-scale task planning. \cite{kim2022representation} explores the representation, learning, and planning algorithms for geometric task and motion planning.

Despite recent advances, TAMP remains a major computational challenge due to its high computational complexity. 
Task planning works with discrete space and is frequently NP-hard \cite{han2018complexity,gaorunning}. Motion planning, on the other hand,  works with continuous space and is known to be PSPACE-hard \cite{hopcroft1984complexity}. Put together, finding feasible solutions to long-horizon TAMP is rather difficult; finding true optimal solutions is generally out of the question. 
%

\textbf{Object Rearrangement.} Object rearrangement is a TAMP challenge with vast applications. NAMO (Navigation among Movable Obstacles) \cite{stilman2005navigation, stilman2008planning}, a class of well-studied problems in the domain, focuses on robot navigation through a space with many movable obstacles that the robot can manipulate. When NAMO instances satisfy certain properties, e.g., monotonicity or linearity, the plan search process can be simplified \cite{okada2004environment, stilman2005navigation, stilman2008planning, levihn2013hierarchical}. 
More recently, rearranging objects on tabletop and shelves  \cite{gao2021fast, cosgun2011push, wang2021uniform, gaorunning, wang2021efficient} have gained attention, in which a robot rearranges multiple objects to pre-defined goal poses. 
In object retrieval \cite{nam2019planning, ZhangLu-RSS-21, vieira2022persistent, huang2022self, huang2021dipn} objects must also be rearranged, but no pre-defined goals for all objects are given. 

Different motion primitives are used for rearrangement tasks, \cite{cosgun2011push, king2017unobservable, vieira2022persistent, huang2021dipn, huang2023optimal} use simple and low-cost but inaccurate non-prehensile actions such as push, some other works \cite{krontiris2015dealing, krontiris2016efficiently, wang2021efficient} use more accurate but higher-cost prehensile actions such as grasp.
Mobile object rearrangement involves more degree of freedoms to move the robot to different place to enrich the workspace \cite{gao2023orla}.

In cases where motion is relatively straightforward to plan, planners \cite{bereg2006lifting, han2018complexity, nam2019planning, gaorunning} often emphasize the task planning challenges inherent in rearrangement tasks. Efficient and effective algorithms \cite{han2018complexity, gaorunning, bereg2006lifting} are developed through converting multi-object rearrangement problems to well-established graph-based discrete optimization problems. Collision-free workspace locations must be allocated for temporary object storage if external free space is unavailable \cite{krontiris2016efficiently,cheong2020relocate,gao2021fast,gao2023utility}. This work adopts ``lazy buffer allocation'' for temporary object storage, which has been shown effective in \cite{gao2021fast} and \cite{gao2022toward} in dual-arm settings.

\textbf{High-DOF Motion Planning.}
Typical manipulators (robot arms) have six degrees of freedom. Planning motion for a $k$-arm system must navigate a very high $6k$-dimensional space.
Computationally, High-dimensional path/motion planning is exceptionally challenging. Typical planners for this purpose fall into two categories: Sampling-based \cite{karaman2011sampling, janson2018deterministic, Ichter2018Learning, gammell2014informed, RRT, PRM} and optimization-based \cite{schulman2014motion, zucker2013chomp, marcucci2023motion, urain2023se, malyuta2022convex}. 
A prevailing approach for high-dimensional motion planning is sampling-based rooted in, e.g., PRM \cite{PRM} and RRT \cite{RRT}, with subsequent and ongoing improvements including RRT*\cite{karaman2011sampling}, SST \cite{li2016asymptotically}, AO-RRT \cite{hauser2016asymptotically}. Recently, VAMP \cite{thomason2023motions} has shown promising results by utilizing AVX2-based parallelization to accelerate forward kinematics and collision-checking computation. 
An optimization-based approach generates trajectories with incremental optimizations based on specific constraints and cost functions. Representative algorithms include CHOMP \cite{zucker2013chomp} and TrajOpt \cite{schulman2014motion}. More recently, \cite{marcucci2023motion} developed convex optimization-based motion planning around obstacles.  \cite{zhang2023provably} implements a Semi-Infinite Program solver to provide provably robust collision avoidance. \cite{urain2023se} studies using diffusion to design smooth joint cost functions for grasp and motion planning. \cite{malyuta2022convex} investigates generating feasible trajectories considering dynamics.
cuRobo \cite{cuRobo} leverages GPU to speed up graph-search for trajectory planning and optimization,  significantly improving planning time and trajectory quality.


\section{Preliminaries}\label{sec:problem}
\subsection{Cooperative Dual-Arm Rearrangement (CDR)}
\begin{figure}[h]
    \centering
    \includegraphics[width=0.23\textwidth]{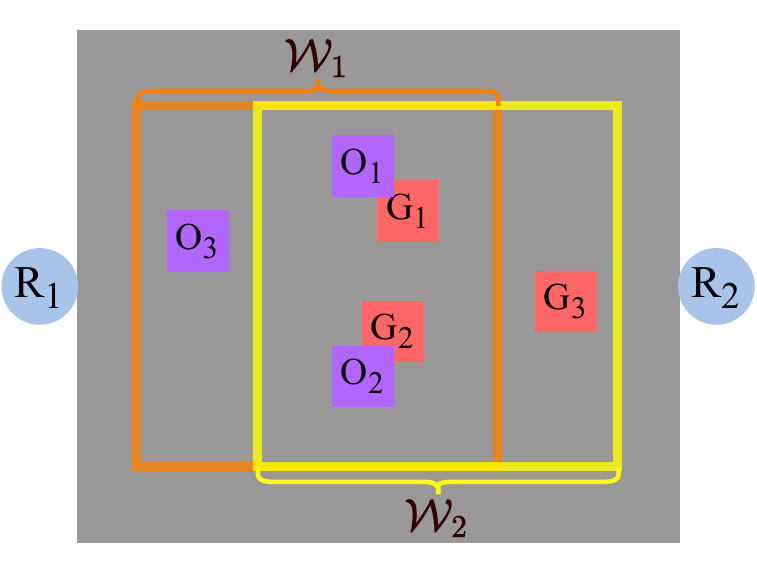} \hspace{2mm}
    \includegraphics[trim={0, 0cm, 0, 0cm}, clip, width=0.23\textwidth]{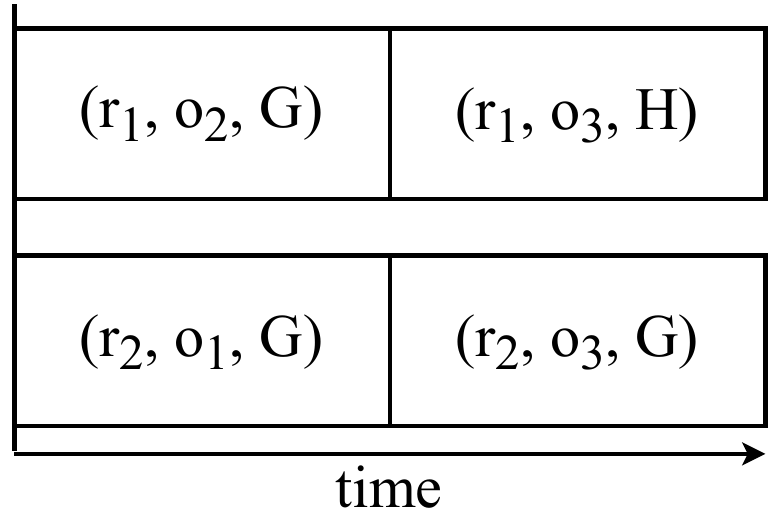}
    \caption{An illustration of the Cooperative Dual-Arm Rearrangement (CDR) setting. [left] Illustrations of the workspace, objects' start/goal poses, and robots' nominal locations. [right] A corresponding task plan with two action steps. 
    }
    \label{fig:task-planning}
    \vspace{-3mm}
\end{figure}

Consider a bounded 2D workspace $\mathcal W$ containing $n$ objects $\mathcal O=\{o_1, o_2, ..., o_n\}$.
Each object $o_i$ has pose $p_i=(x_i,y_i,\theta_i)\in SE(2)$.
An arrangement $\mathcal A=(p_1,p_2,...,p_n)$ is \emph{feasible} if (1) objects are not in collision, and (2) each object is entirely contained in $\mathcal W$. 
The objects are rearranged using two robot arms $r_j, j = 1, 2$,
on opposite sides of $\mathcal W$. Due to reachability, robot $r_j$  only serves  $\mathcal W_j \subset \mathcal W$, which naturally partition $\mathcal W$ into three regions shown in Fig.~\ref{fig:task-planning}. We define the overlap ratio of $\mathcal W$ as $\rho:=(\mathcal W_1 \bigcap \mathcal W_2)/(\mathcal W_1 \bigcup \mathcal W_2)$.

The setup mirrors practical settings in which a robot has limited reachability, requiring two operation modes:
%
\begin{itemize}[leftmargin=4mm]
\item When an object's start/goal poses are in the same $\mathcal W_j$, $j = 1, 2$, and when the goal pose is free, a robot can use \emph{overhand pick-n-place} to rearrange the object. When executing an overhand pick/place, a robot moves to a pre-picking/placing pose, which is $k$cm above the picking/placing pose, goes down to the picking/placing pose in a straight line, picks/places the object, and finally moves back to the pre-picking/placing pose in a straight line.
\item Otherwise, a \emph{handoff} between the two robots must be performed to complete the rearrangement of the object. 
In a handoff, each robot moves to a pre-handoff pose and then moves to the handoff pose in a straight line, executing the handoff. After the handoff, they return to the pre-handoff poses before changing the orientation.
\end{itemize}
During the rearrangement, the configuration of a robot $r_i$ with $n$ joints and one \textit{close/open} gripper is represented as $q^i:=\{j_1^i, j_2^i, ..., j_n^i, g^i\}$, where $j_k^i$ is the $k^{th}$ joint angle of $R_i$ and $g^i\in \{0,1\}$ indicates whether the gripper of $r_i$ is open ($1$) or closed ($0$).
In our implementation, we assume the gripper is closed if and only if an object is in hand.
Therefore, we only consider the first $n$ degree of freedom for trajectory planning.
In the rearrangement, a plan can be represented as a sequence of collision-free dual-arm configurations $\tau = \{q_t | t=0, \cdots, n\}$ where $q_t:=(q_t^1, q_t^2)$ is the planned configuration of the $t^{th}$ waypoint and $n$ is the length of the trajectory.

In summary, we study the problem of \emph{Cooperative Dual-Arm Rearrangement} (CDR): Given feasible start and goal arrangements of workspace objects $\mathcal A_s$ and $\mathcal A_g$, seek a dual-arm rearrangement plan $\tau$.
\vspace{-3mm}

\subsection{cuRobo-Based Motion Generation}

\ours leverages cuRobo \cite{sundaralingam2023cuRobo}, a GPU-based efficient trajectory planner and motion optimizer, to assist our path and motion planning effort. 
Given start/goal configurations, cuRobo computes feasible and smooth trajectories by first computing inverse kinematics (IK) solutions for the goal. Then, it finds collision-free \emph{seed trajectories} from the start state to the IK solutions using a parallel geometric planner. The initial trajectories are then optimized to be smooth, more time-efficient while conforming to robot dynamics.
cuRobo provides magnitudes faster trajectory computation/optimization in high-dimensional spaces, including supporting planning for dual-arm systems.
However, it can only plan motions synchronously. In other words, the two robots must reach their respective goals simultaneously, which does not directly apply to cooperative rearrangement tasks. 
cuRobo also frequently produces smooth but long trajectories with undesired goal IK solutions.
In this work, we introduce strategies to maximize cuRobo's strengths and mitigate its limitations for optimal performance in the dual-arm TORO problem.


\section{Makespan-Optimized Dual-Arm Planning}\label{sec:algorithms}
\subsection{Task Planning Phase}
\ours first compute a base, synchronous task plan largely following the approach of \emph{Manipulation Cost-based makespan Heuristic Search} (MCHS) \cite{gao2022toward}. MCHS minimizes the number of \emph{action steps} in the rearrangement plan. At each action step, a robot arm can complete an individual pick-n-place, a coordinated handoff, or idle.
Fig.~\ref{fig:task-planning} shows a rearrangement example and its corresponding two-step task plan computed by MCHS.
In the first step, $R_1$ and $R_2$ pick up $o_2$ and $o_1$ and place them at their goals, respectively.
In the second step, $R_1$ passes $o_3$ to $R_2$, then $R_2$ places $o_3$ at its goal pose.

MCHS does not consider the motion planning for the dual-arm system. In reality, robot arms take different times to rearrange a single object and may need to coordinate to avoid collisions in the shared workspace. 
Using the task plan produced by MCHS, we decouple primitive actions in the plan into lower-level \emph{sub-tasks}.
For example, for the primitive action $(r_1, o_3, H)$ in Fig.~\ref{fig:task-planning}, it can be decoupled into four sub-tasks of $R_1$: move to pre-handoff pose, move to handoff pose, open gripper, and move back to pre-handoff pose.
Even though the original task plan is synchronized (i.e., two robots start and finish individual actions simultaneously in each action step), the generated sub-task sequences allow us to enable asynchronous execution and generate smooth and executable dual-arm motions.
The sub-tasks are categorized as follows.
\begin{enumerate}[leftmargin=5mm]
    \item A \textbf{Traveling Task (TT)} includes motions between home, pre-picking, pre-placing, and pre-handoff poses. 
    A TT sub-task can be represented as $(p_1, p_2)$, where $p_1,p_2\in SE(3)$, and the robot needs to move from the current pose $p_1$ to the target pose $p_2$.
    In our implementation, we also treat idling as a TT sub-task where $p_1=p_2$.
    \item A \textbf{Manipulation Task (MT)} includes motions between a picking/placing/handoff pose and its corresponding pre-picking/pre-placing/pre-handoff pose.
    An MT sub-task can be represented as $(p_1, p_2)$, where $p_1,p_2\in SE(3)$, and the robot needs to move in a straight line from the current pose $p_1$ to the target pose $p_2$.
    \item A \textbf{Gripper task (GT)} includes opening/closing grippers when picking/placing an object or executing a handoff. 
    A GT task can be represented as $(p_1,k)$, where the robot arm stays at the current pose $p_1$ for $k$ timesteps.
\end{enumerate}


\subsection{Motion Planning Phase}
\ours's dual-arm motion planning framework is described in Algo.\ref{alg:MotionPlanning}.
Each robot executes its sub-tasks sequentially; trajectory conflicts are resolved by leveraging cuRobo's motion generator.
In Line 1, we initialize $\mathcal T$, $\pi_1$, and $\pi_2$, where $\mathcal T$ is the dual-arm trajectory, $\pi_1$ and $\pi_2$ are current sub-tasks of $R_1$ and $R_2$. 
The rearrangement is finished if all of $\Pi_1, \Pi_2, \pi_1, \pi_2$ are empty (Line 2). 
In Line 3, $\pi_1$ and $\pi_2$ are updated.
Specifically, if the current sub-task is finished, the next sub-task in the sequence kicks in. 
Given $\pi_1, \pi_2$, single-arm trajectories $\tau_1, \tau_2$ are computed (Line 4-5). 
$\tau_1$ and $\tau_2$ avoid self-collisions and collisions between the planning robot arm and the workspace, but they don't consider collisions between arms.
If there is no collision between $\tau_1$ and $\tau_2$, a dual-arm trajectory is generated by merging $\tau_1$ and $\tau_2$ until one of the trajectories reaches its ending timestep (Line 7).
Otherwise, there are two cases (Line 8 - 12): (1) both trajectories can be replanned (Line 8-10); (2) only one trajectory can be replanned (Line 11-12).
For (1), we sample a collision-free dual-arm goal configuration $q^\star$(Line 9). At $q^\star$, one robot finishes this sub-task while the other is on its way.
And then, we compute dual-arm trajectory via cuRobo (Line 10) from the current dual-configuration to $q^\star$.
For (2), we compute a dual-arm trajectory that prioritizes the sub-task of one arm, and the other arm yields as needed. (Line 11-12).
The details of our planner for the two cases are described in Sec.\ref{sec:dual_planning}.
A third case beyond (1) and (2) exists in theory where neither of the trajectories can be replanned. We avoid this case by providing additional clearance for collision checking.
A small clearance is enough since non-TT sub-task trajectories are short (a few centimeters).
The planner computes a dual-arm trajectory $\tau$, sped up by Toppra\cite{pham2018new}, a time-optimal path parametrization solver, and added to $\mathcal T$ (Line 13).
\begin{algorithm}
\begin{small}
    \SetKwInOut{Input}{Input}
    \SetKwInOut{Output}{Output}
    \SetKwComment{Comment}{\% }{}
    \caption{Fast Dual-Arm Motion Planning}
		\label{alg:MotionPlanning}
    \SetAlgoLined
		\vspace{0.5mm}
    \Input{$\{\Pi_1, \Pi_2\}$: sub-task sequences
    }
    \Output{$\mathcal T$: dual-arm motion}
\vspace{0.5mm}

$\mathcal T\leftarrow \emptyset$; 
$\pi_1, \pi_2\leftarrow \emptyset, \emptyset$;\\
\While{
not Finished($\Pi_1, \Pi_2, \pi_1, \pi_2$)
}{
$\pi_1, \pi_2\leftarrow $UpdateSubTasks($\Pi_1, \Pi_2, \pi_1, \pi_2$);\\
$\tau_1\leftarrow SingleArmPlanning(\pi_1)$;\\
$\tau_2\leftarrow SingleArmPlanning(\pi_2)$;\\
\If{not $InCollision(\tau_1,\tau_2)$}{
    $\tau\leftarrow$ MergeTraj($\tau_1,\tau_2$);\\
}
\ElseIf{$\pi_1$ is TT and $\pi_1$ is TT}{
$q^\star\leftarrow $ SampleDualGoal($\tau_1,\tau_2$);\\
$\tau \leftarrow$ DualArmPlanning($q^\star$);\\
}
\lElseIf{$\pi_1$ not TT}
{
$\tau\leftarrow$ PlanPartialPath($\tau_1$)
}
\lElseIf{$\pi_2$ not TT}
{
$\tau\leftarrow$ PartrialPlanning($\tau_2$)
}
$\tau\leftarrow$ ToppraImprovement($\tau$);
$\mathcal T \leftarrow \mathcal T + \tau$\\
}
\Return $\mathcal T$;\\
\end{small}
\end{algorithm}
\vspace{-6mm}

\subsection{Enhancing Inverse Kinematic (IK) Solutions}\label{sec:single_planning}
Given the goal end-effector pose $p_2$, cuRobo motion generator will sample IK seeds and compute multiple IK solutions of $p_2$.
However, it may lead to undesired IK solutions of $p_2$.
For example, in IK computation for pre-picking poses, cuRobo may select a solution where the elbow is positioned below the end effector, preventing the robot from executing the subsequent grasping.
We overcame issues caused by undesirable IK solutions by dictating IK seeds. The goal pose's seed is set to the current configuration for MT tasks because we want the target configuration to be close.
The current configuration may be too far from the target configuration for TT tasks, leading to the undesired configuration mentioned above.
Therefore, in TT tasks, the seed is set to a specific configuration with its corresponding end-effector pose $30$ cm above its reachable workspace region as shown in Fig.~\ref{fig:simnreal-setup}.
\subsection{Improving Path Planning to Avoid Detours}
cuRobo's geometric (path) planner samples graph nodes around the straight line between the start, goal, and home (retract) configurations. Based on these nodes, seed trajectories are planned via graph search. 
When the planning problem is challenging (e.g., the second arm blocks the straight line between the start and goal configurations for the first arm), the computed trajectories tend to make lengthy detours by detouring through the neighborhood of the home configuration, which wastes time and yields unnatural robot paths. This is especially problematic when solving many rearrangement sub-tasks. 
We improve seed trajectory quality for each arm by uniformly sampling 32 overhand poses above the workspace. 
The corresponding IK solutions of these manipulation-related poses form 1024 dual-arm configurations, among which the collision-free configurations are added to the geometric graph. This approach proves to be highly effective in avoiding unnecessary detours.

\subsection{Strategies to Resolve Dual-Arm Trajectory Conflicts}\label{sec:dual_planning}
Two arms must coordinate to resolve conflicts and progress toward their next sub-tasks when collisions exist between individual trajectories.
As mentioned in Algo.~\ref{alg:MotionPlanning}, there are two different collision scenarios.
In the first, when both robots have TT sub-tasks, neither has a specific trajectory to follow.
In this scenario, we sample a collision-free dual-arm configuration (Algo.~\ref{alg:sample_dual_goal}) and plan a trajectory from the current configuration to the sampled configuration.
In the second, when one of the robots has a specific trajectory to follow, the other arm needs to plan a trajectory to avoid collision.
Since cuRobo does not support planning among dynamic obstacles, we propose an algorithm, PlanPartialPath (Algo.~\ref{alg:partial}), to resolve conflicts.
The details of the algorithms are elaborated on in the following sections.

\subsubsection{Dual-arm goal sampling}
Algo.~\ref{alg:sample_dual_goal} takes single arm trajectories $\tau_1$ and $\tau_2$ as input and outputs the collision-free dual-arm configuration $q$.
Without loss of generality, we assume $\tau_1$ is the shorter trajectory in time dimension.
Usually, this means that the sub-task goal configuration of $R_1$ is closer to the current configuration.
Therefore, we sample a configuration as the goal configuration to the dual-arm motion planning problem, where $R_1$ reaches the sub-task goal configuration while $R_2$ progresses toward its goal configuration.
In Lines 1-2, we let $q_1^\star$ be the sub-task goal configuration of $R_1$ and $q_2^\star$ be the configuration in $\tau_2$ when $R_1$ reaches $q_1^\star$.
If the dual-arm configuration $(q_1^\star,q_2^\star)$ is collision-free, then we output the configuration (Line 3).
Otherwise, we sample the configuration of $R_2$ in the neighborhood of $q_2^\star$ (Line 5-11). 
The sampling starts with the $\gamma_0-$sphere centered at $q_2^\star$ and the range is constantly expanded to a $\gamma_n-$sphere, where $\gamma_0$ and $\gamma_n$ are constants.
When local sampling fails, 
we let $R_2$ yield to $R_1$ by sampling $R_2$ configurations on the straight line between its current configuration and its home configuration since the home configuration is guaranteed to be collision-free with the other arm.
Let $S_2$ be the discretization (with $k_2$ waypoints and $k_2$ is a constant) of the straight-line trajectory between the current configuration of $R_2$ and the home configuration of $R_2$ (Line 12).
We then find the collision-free waypoint of $R_2$ nearest the current configuration and return the corresponding dual-arm configuration (Line 13-15).
\vspace{-3mm}

\begin{algorithm}
\begin{small}
    \SetKwInOut{Input}{Input}
    \SetKwInOut{Output}{Output}
    \SetKwComment{Comment}{\% }{}
    \caption{SampleDualGoal}
		\label{alg:sample_dual_goal}
    \SetAlgoLined
		\vspace{0.5mm}
    \Input{$\tau_1:$ shorter single-arm trajectory, $\tau_2:$ longer single-arm trajectory, $k_1, k_2, \gamma_0, \gamma_n:$ hyper-parameters
    }
    \Output{$q$: collision-free dual-arm configuration}
\vspace{0.5mm}

$q^\star_1\leftarrow$ Ending configuration of $\tau_1$;\\
$q^\star_2\leftarrow$ configuration of $\tau_2$ when $\tau_1$ ends;\\
\lIf{CollisionFree($q^\star_1, q^\star_2$)}
{
\Return Merge($q^\star_1,q^\star_2$);
}
\Else{
$r = \gamma_0$;\\
\While{$r<\gamma_n$}{
    \For{$i=1$ to $k_1$}
    {
    $q_2\leftarrow$ Sample configuration in the $r$-sphere of $q^\star_2$;\\
    \If{CollisionFree($q^\star_1, q_2$)}
    {
    \Return Merge($q^\star_1,q_2$)
    }
    }
    $r\leftarrow 2r$;\\
}
$S_2\leftarrow k_2$-discretization of the straight line between the start of $\tau_2$ and robot home configuration;\\ 
\For{$q_2\in S_2$}
{
\If{CollisionFree($q^\star_1, q_2$)}
    {
    \Return Merge($q^\star_1,q_2$)
    }
}
}
\end{small}
\end{algorithm}

\vspace{-3mm}
\subsubsection{Priority based planning}
In Algo.~\ref{alg:partial}, we compute a collision-free dual-arm trajectory when one of the robots has a pre-defined trajectory to follow.
Without loss of generality, we assume $R_1$ has a pre-defined trajectory $\tau_1$ to execute while $R_2$ needs to yield.
We first compute a collision-free trajectory $\tau_2$ for $R_2$ from the current configuration to its home configuration assuming $R_1$ is fixed at the current configuration (Line 2).
In the worst case, it allows $R_1$ to execute its pre-defined trajectory $\tau_1$ after $R_2$ moves back to its home configuration.
In Line 3-9, we progress on executing $\tau_1$ and $R_2$ yields along $\tau_2$ as needed.
$\tau_1[i_1]$ is the next configuration of $R_1$ on $\tau$.
In Line 5, we compute the first index $i_2'$ such that $i_2'\geq i_2$ and $\tau_2[i_2':]$ is collision-free with $\tau_1[i_1]$.
Therefore, $\tau_2[i_2], ..., \tau_2[i_2'-1]$ is collision-free with $\tau_1[i_1-1]$, and $\tau_2[i_2']$ is collision-free with $\tau_1[i_1]$.
We add the collision-free dual-arm configurations to $\tau$ (Line 6-7) and update indices $i_1$ and $i_2$ (Line 8).
To summarize, if $R_2$ blocks $\tau_1$, then $R_2$ yields along $\tau_2$ as needed; otherwise, $R_2$ stays.

\begin{algorithm}
\begin{small}
    \SetKwInOut{Input}{Input}
    \SetKwInOut{Output}{Output}
    \SetKwComment{Comment}{\% }{}
    \caption{PlanPartialPath}
		\label{alg:partial}
    \SetAlgoLined
		\vspace{0.5mm}
    \Input{$\tau_1:$ pre-defined trajectory
    }
    \Output{$\tau$: dual arm trajectory}
\vspace{0.5mm}
$\tau\leftarrow \emptyset;$\\ 
$\tau_2\leftarrow$ Motion planning between the current configuration and home configuration of $R_2$ while $R_1$ stays at the current configuration;\\
$i_1, i_2\leftarrow 0, 0$;\\
\While{$i_1<|\tau_1|$}
{
$i_2'\leftarrow$ CollisionFreeFirstIndx($\tau_1[i_1], \tau_2[i_2:]$);\\
$\tau\leftarrow$ $\tau$.Extend($\tau_1[i_1-1],[\tau_2[i_2],\dots, \tau_2[i_2'-1]]$);\\
$\tau\leftarrow$ $\tau$.Extend($\tau_1[i_1],[\tau_2[i_2']]$);\\
$i_1+=1$; $i_2=i_2';$
}
\Return $\tau$;
\end{small}
\end{algorithm}


\section{Adaption to real robot system}\label{sec:real}
\subsection{Real2Sim2Real}
Solving CDR mainly involves computing solutions, simulating them, and transferring them to real robot systems. To apply \ours to real robot systems, however, requires us first to build an accurate model (digital twin) of the actual system.
To achieve this, we calibrate the robots relative to each other by placing them (at different times) at the same seven diverse positions in the workspace alternately. This allows us to compute an accurate transformation between the two robot arms, upon which a digital twin of the actual setup is obtained. 
All the objects to be manipulated are labeled with ArUco markers\cite{garrido2014automatic}, and we can directly transfer their positions into the simulation with a calibrated camera.

\begin{figure}[h]
    \vspace{-1mm}
    \centering
\includegraphics[width=1\columnwidth]{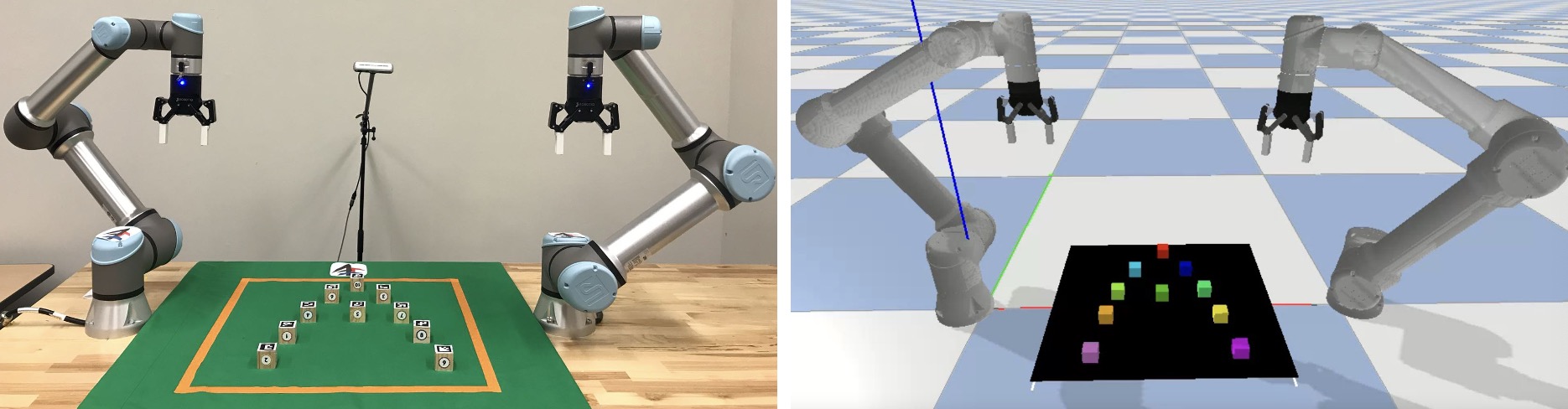}
    \caption{[left] A configuration of our dual-arm setup, also showing that the camera is mounted at the side to detect objects' poses. [right] The corresponding Dual-Arm setup in pyBullet simulation.}
    \label{fig:simnreal-setup}
    \vspace{-3mm}
\end{figure}

Since the two real robot arms are calibrated and their positions are transferred into the simulation, the trajectories planned in simulations can be seamlessly transferred to the real robot system and executed directly without further post-processing. Fig.~\ref{fig:simnreal-setup} shows that our real and simulated setups mirror each other.

\subsection{Real Robot Control}
The open-source Python library ur\_rtde\cite{urrtde} is used to communicate with and control the robot, which allows us to receive robot pose/configuration and send control signals in real-time (up to 500Hz). Our controller sends joint positions for continuous control via the ServoJ() function, a PD-styled controller. The control signals are sent out alternatively at every cycle (e.g., with a 0.002-second interval), and synchronized between the two arms.

\section{Experimental Studies}\label{sec:evaluation}
In this section, we evaluate our \ours and compare it with two baseline planners both in simulation and on real robots.
The two baseline planners are:
\begin{enumerate}[leftmargin=5mm]
    \item \textbf{Baseline (BL)} Similar to \ours, BL computes collision-free dual-arm motion plans based on MCHS task plans.
    When the clearance between robots is larger than a certain threshold, both arms execute their tasks individually.
    Otherwise, one arm yields by moving toward its home state, which is guaranteed to be collision-free.
    After obtaining a collision-free initial trajectory, BL smoothens the yielding trajectories: instead of switching frequently between yielding and moving toward the next target, the robot attempts to move back to a collision-free state and waits until the conflict is resolved.
    \item \textbf{Baseline+Toppra (BL-TP)} BL-TP uses Toppra to speed up the trajectory planned by BL.
\end{enumerate}

\subsection{Sim2Real Gap: Trajectory Tracking Accuracy}
\begin{figure}[h]
    \centering
        \includegraphics[width=0.95\columnwidth]{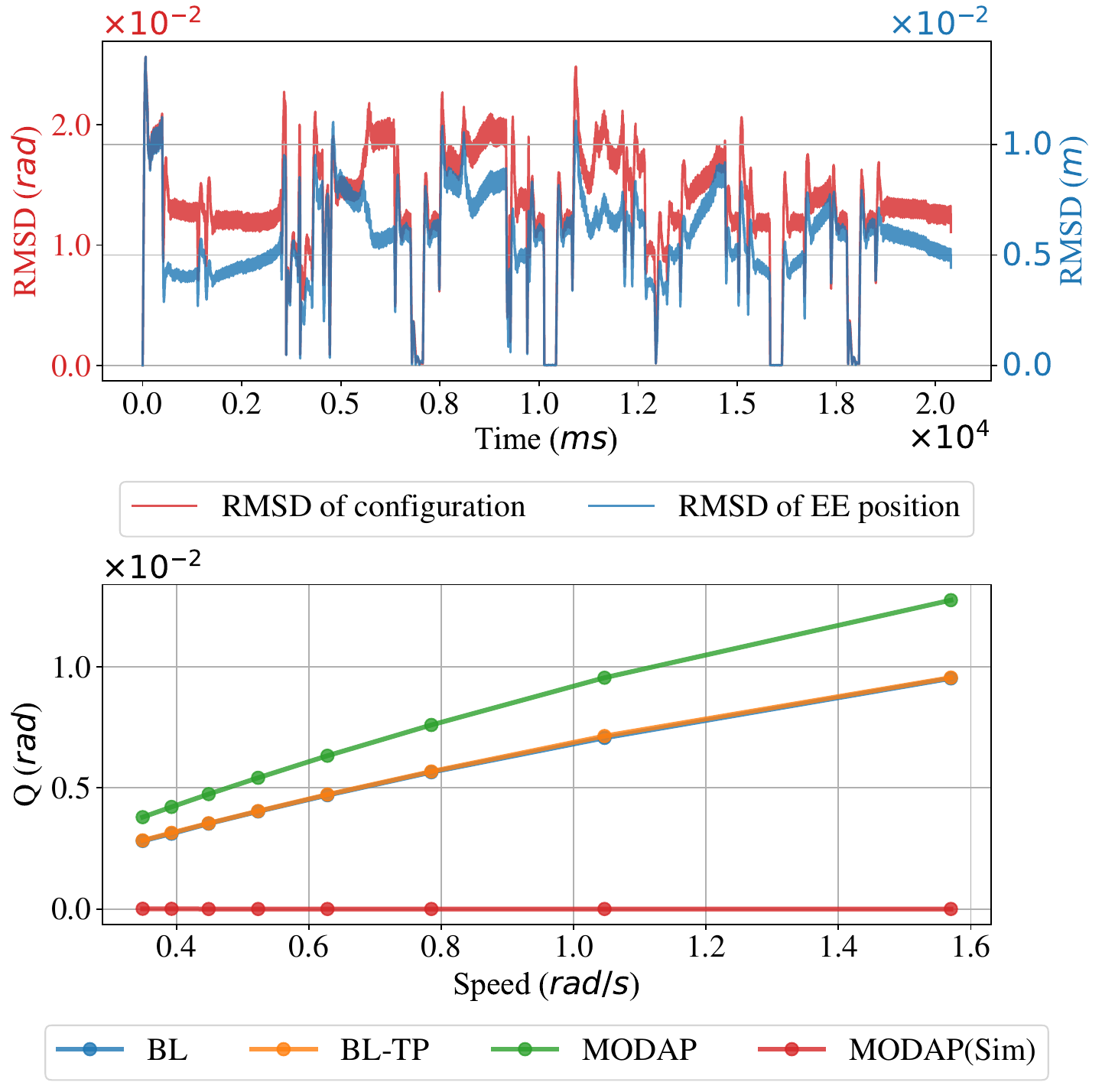}
    \caption{[top] RMSD of configuration and end-effector position on a sampled trajectory executed with a maximum speed of 1.57rad/s on real robots. [bottom] The average Q value of 5 sampled trajectories executed at different maximum speeds (sim denotes execution in simulation; otherwise, the execution is on a real robot). Note that the lines of BL and BL-TP overlap with each other.} 
    \label{fig:real_data}
    \vspace{-6mm}
\end{figure}

Whereas we must evaluate plans on real robots to ensure they work in the real world, doing so extensively prevents us from running a large number of experiments as it is time-consuming to set up and run real robots. 
To that end, we first evaluate the sim2real gap of our system. 
The Root-Mean-Square Deviation(RMSD) is introduced as a metric to evaluate the accuracy of the real robot control for one waypoint $q_i$ in the planned trajectory $\tau$
\vspace{-2mm}
\begin{equation*}
    RMSD(q_i) = \sqrt{\frac{\left( \hat{q_i} - q_i\right)^T \left( \hat{q_i} - q_i\right)}{\left| q_i\right|}}
\vspace{-2mm}
\end{equation*}
where $\hat{q_i}$ denotes the real configuration retrieved from the robot while executing the control signal $q_i$ and $\left| q_i \right|$ denotes the degree of freedom of the robot system. The control quality of a trajectory can be evaluated as
\vspace{-2mm}
\begin{equation*}
     Q(\tau) = \mathbb{E}[RMSD(q_i)]
\vspace{-2mm}
\end{equation*}
Trajectories are collected from 5 different scenes with four randomly placed cubes as the start and goal, planned by different methods, and then executed in simulation and real-world at different speed limits. 
In the meantime, RMSD and Q for the end-effector position can be recorded.
In Fig.~\ref{fig:real_data}[top], we show the RMSD of the configuration and the end effector's position on each waypoint in a trajectory planned using \ours. 
We observe that we can see that the execution error is about $1^{\circ}$ for each joint and 5mm as the end-effector position is concerned, executing at a speed of 1.57rad/s at every joint.
\begin{figure}[ht]
    \centering
\includegraphics[width=0.98\columnwidth]{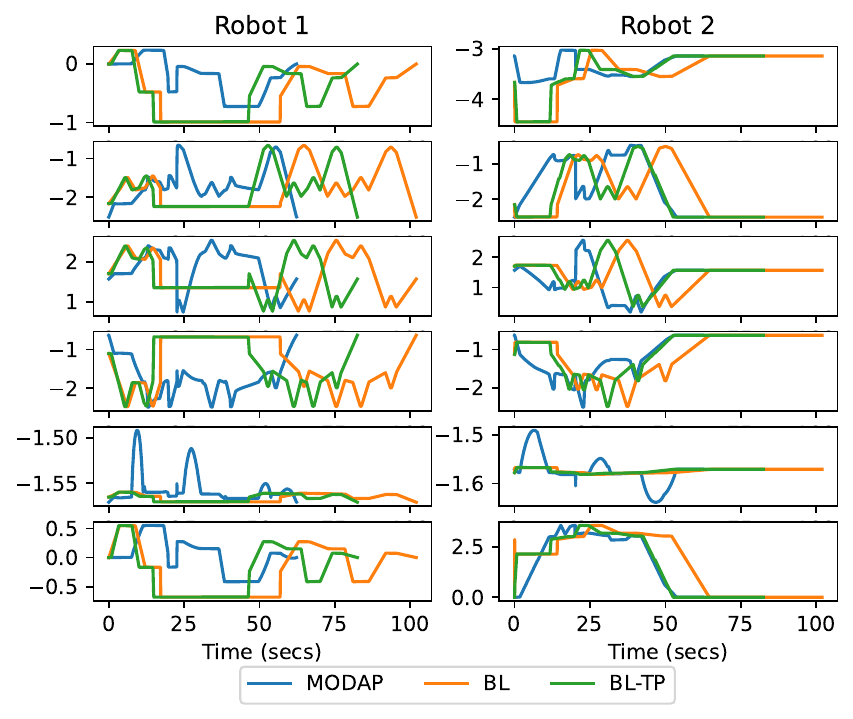}
    \vspace{-2mm}
    \caption{Trajectories of BL, BL-TP, and \ours for the a typical rearrangement task. Each row, from top to bottom, corresponds to a robot joint in radians from one to six. Each column shows the trajectory of the six joint angles of a robot.}
    \label{fig:sim_exec_joints}
\end{figure}
Fig.~\ref{fig:real_data}[bottom] shows the average configuration Q value of 5 trajectories executed at different speed limits. 
First, we note that trajectory execution is perfect in simulation, as expected. 
We observed that \ours induces a slightly larger (though perfectly acceptable) execution error than BL and BL-TP. This is due to trajectories produced by \ours running at faster overall speeds. 

Overall, our evaluation of the trajectory tracking accuracy indicates that the sim2real gap is negligibly small across the methods, including \ours. This suggests we can confidently plan in simulation and expect the planned time-parameterized trajectory to execute as expected on the real system (see our video for a side-by-side execution comparison). It also means that we can compare the optimality of the methods using simulation, which we perform next. 

\subsection{Performance on Rearrangement Plans Execution} 
In evaluating the performance of \ours and comparing that with the baseline methods, we first examine the execution of \ours, BL, and BL-TP on a typical instance involving the rearrangement of five objects with $\rho=1.0$ (full workspace overlap). The time-parameterized trajectories of the six robot joints are plotted in Fig.~\ref{fig:sim_exec_joints} for all three methods.
We can readily observe that \ours is much more time-efficient than BL and BL-TP. 
We can also readily observe the reason behind this: 
BL has one arm idling in most of the execution due to inefficient dual-arm coordination. 
After Toppra acceleration, BL-TP is still $32\%$ slower than \ours.

\begin{figure}[ht]
\vspace{-1.5mm}
    \centering
\includegraphics[width=0.48\columnwidth]{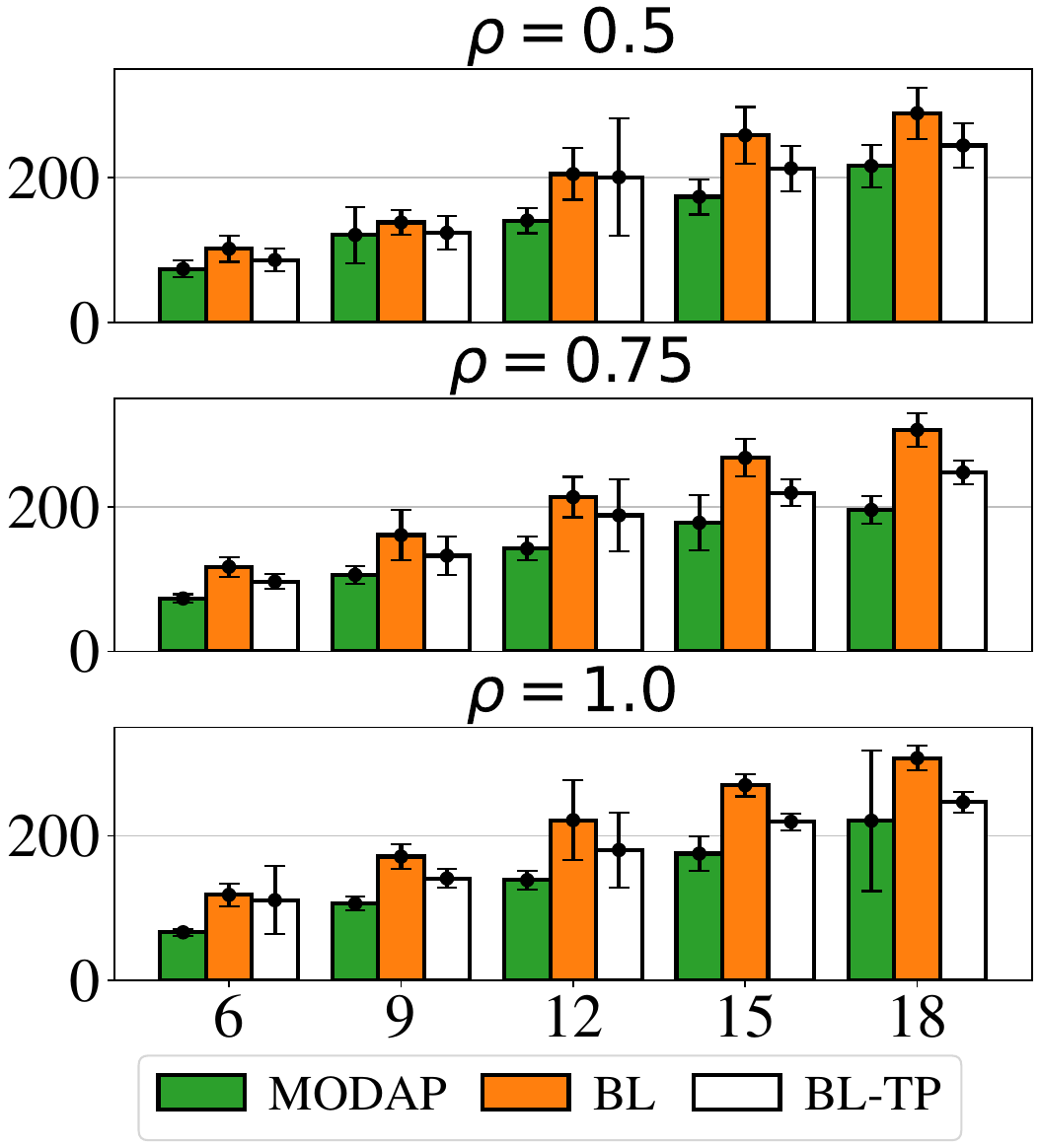}
\includegraphics[width=0.48\columnwidth]{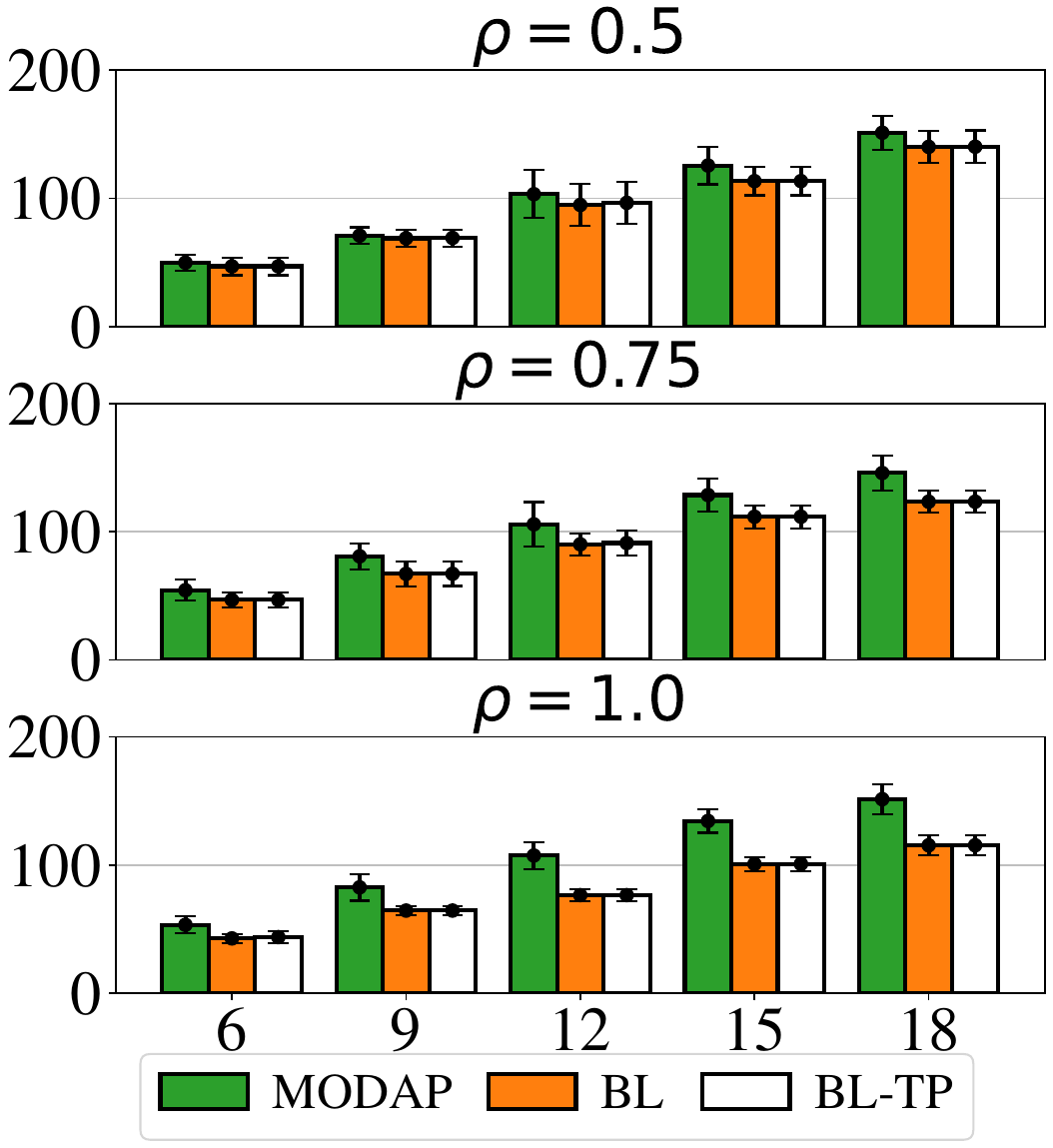}
    \caption{Average execution time (left) and trajectory total length (right) computed by compared methods under different overlap ratios $\rho$ and the number of workspace objects.}
    \label{fig:sim_data}
\vspace{-3mm}
\end{figure}

\begin{figure}[ht]
    \centering
\includegraphics[trim={48cm, 22cm, 48cm, 25cm}, clip, width=0.24\columnwidth]{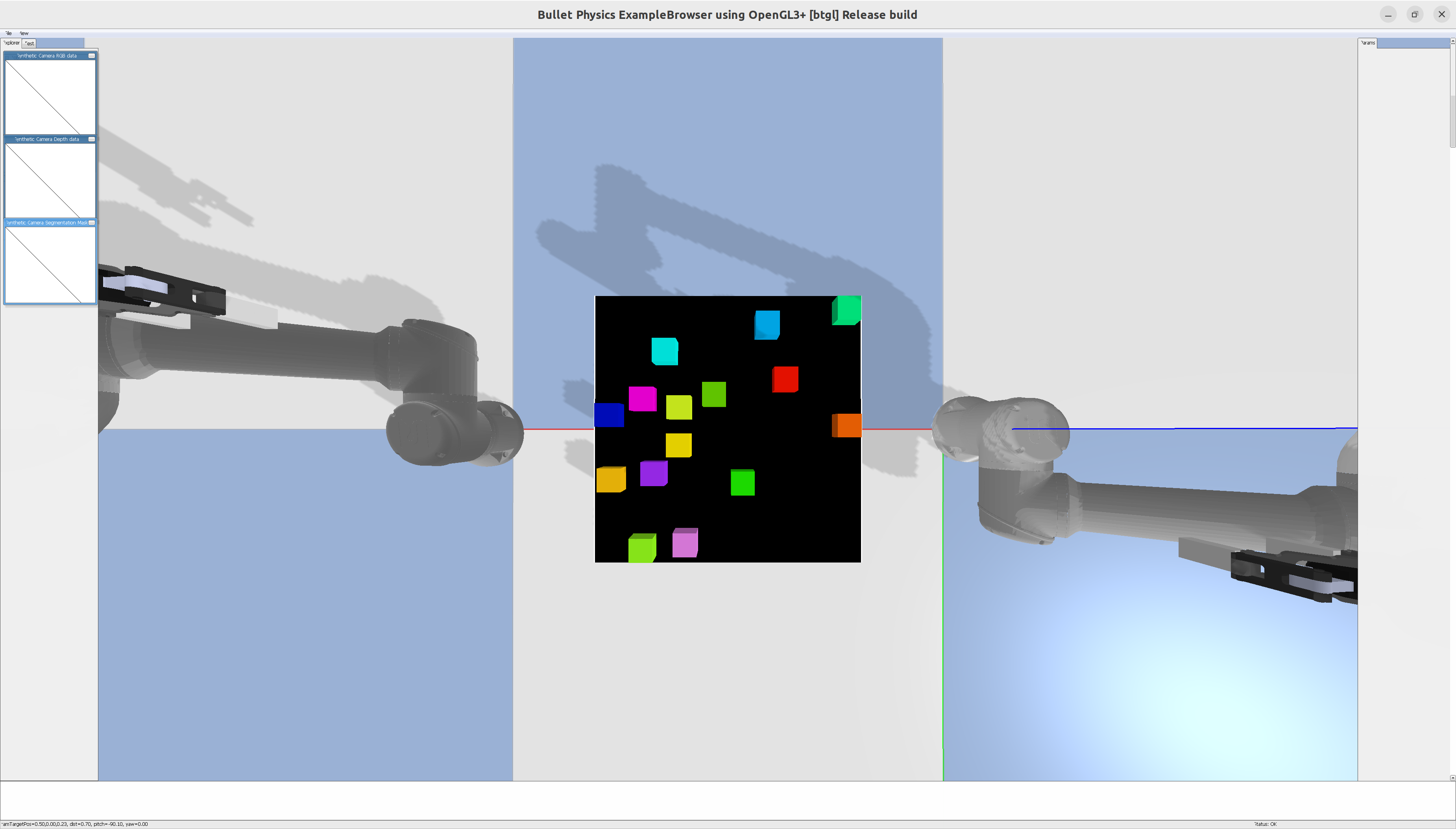}
\put(-35, -10){(a)}
\includegraphics[trim={48cm, 22cm, 48cm, 25cm}, clip, width=0.24\columnwidth]{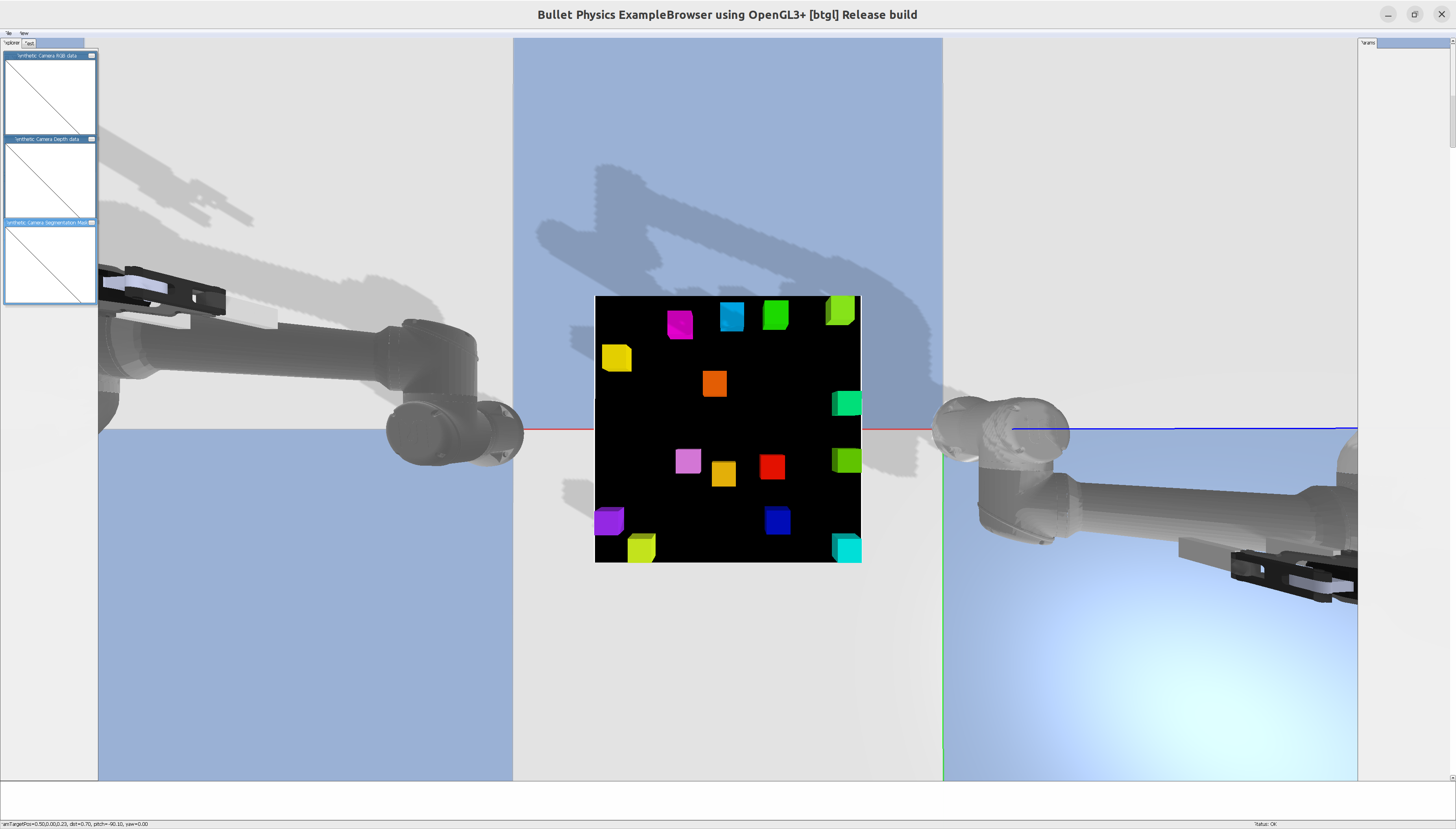}
\put(-35, -10){(b)}
\includegraphics[trim={48cm, 22cm, 48cm, 25cm}, clip, width=0.24\columnwidth]{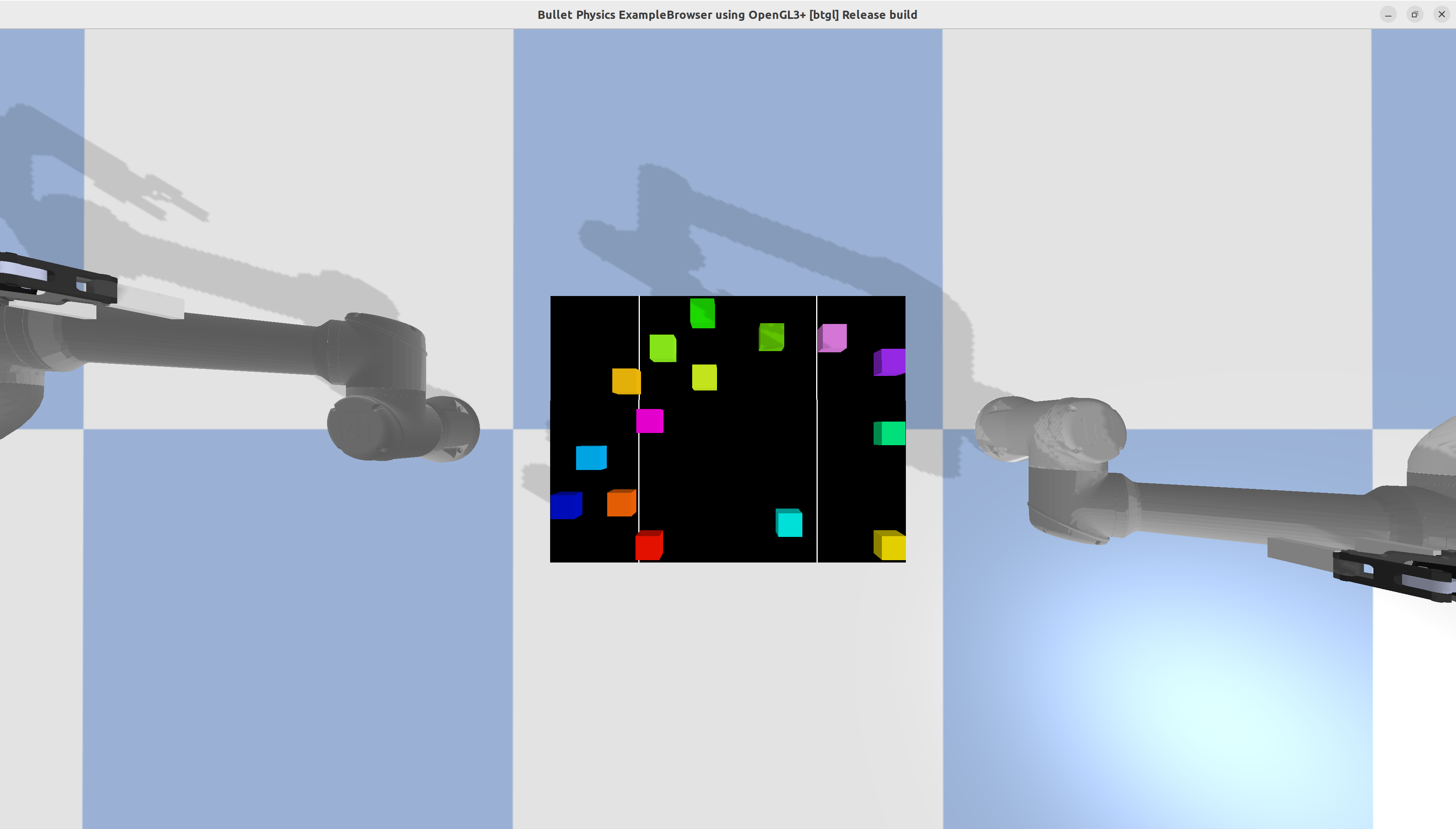}
\put(-35, -10){(c)}
\includegraphics[trim={48cm, 22cm, 48cm, 25cm}, clip, width=0.24\columnwidth]{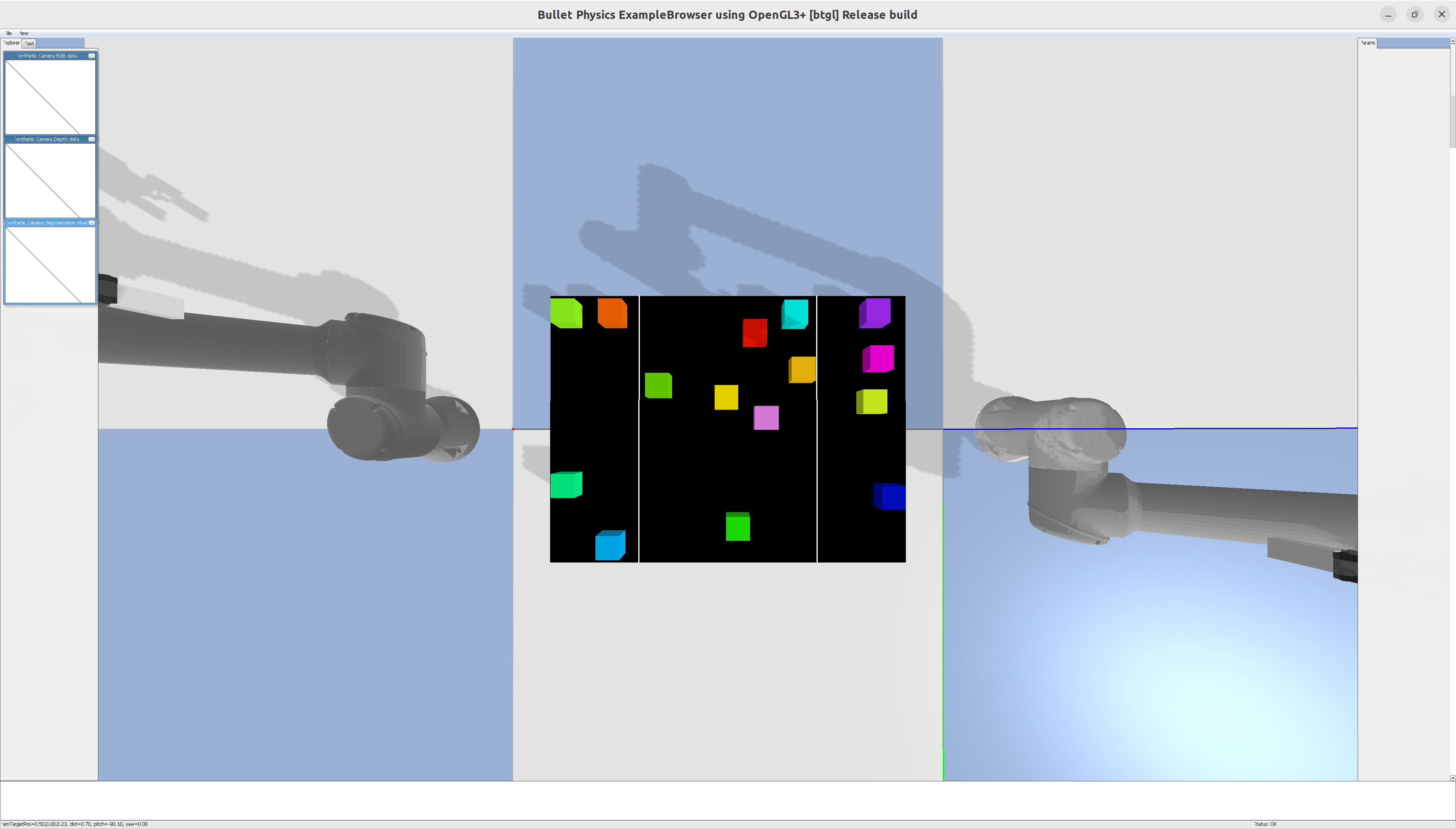}
\put(-35, -10){(d)}
    \caption{Two examples of start and goal configurations with 15 objects at overlap rates $\rho=1$ (a, b) and $\rho=0.5$ (c, d). (a) and (c) (resp., (b) and (d)) are starting (resp., goal) configurations.}
    \label{fig:sim_workspace}
\vspace{-3mm}
\end{figure}

Fig.~\ref{fig:sim_data} shows the result of a full-scale performance evaluation under three overlapping ratios ($\rho = 0.5, 0.75, 1$) and different numbers of objects to be rearranged ranging between $6$-$18$. 
The simulation experiments are executed on an Intel$^\circledR$ Core(TM) i7-9700K CPU at 3.60GHz and an NVIDIA GeForce RTX 2080 Ti GPU for cuRobo planner. For each method, the average execution time and total trajectory length are given; each data point (with both mean and standard deviation) is computed over $20$ randomly generated instances. Two example test cases are shown in  Fig.~\ref{fig:sim_workspace}. We set the maximum per-instance computation time allowance to be 500 seconds, sufficient for all computations to complete. \ours, exploring many more trajectories than BL and BL-TP, takes, on average, 2-3 times more computation time. 
The total length is the sum of robot trajectory lengths computed under the L2 norm in the 6-DoF robot's configuration space.
We observe that \ours's plans can be executed faster than BL and BL-TP even though cuRobo generates longer trajectories, which aligns with the observation based on Fig.~\ref{fig:sim_exec_joints}. 
When $\rho=0.5$, \ours saves $13\%-32\%$ execution time compared with BL, and $3\%-30\%$ compared with BL-TP.
When $\rho=1.0$, \ours saves $29\%-44\%$ execution time compared with BL, and $11\%-40\%$ compared with BL-TP.
\ours shows apparent efficiency gain as the overlap ratio $\rho$ increases, which shows that the proposed \ours effectively resolves conflicts between robot trajectories.

\section{Conclusion}\label{sec:conclusion}
In conclusion, we study the joint optimization of task and motion planning for tabletop object rearrangement with a dual-arm setup, yielding the design of an efficient planner \ours that takes in a tabletop rearrangement task and outputs the planned trajectory while conforming to specified robot dynamics constraints, e.g., velocity, acceleration, and jerk limits. Compared to the previous state of the art, \ours achieve up to $40\%$ execution time improvement even though the planned trajectories are longer. In particular, \ours shines in cases where extensive joint dual-arm planning must be carried out to compact the overall task and motion plan. 

{\small
\bibliographystyle{formatting/IEEEtran}
\bibliography{bib/bib}
}


\end{document}